# Development of Word Embeddings for Uzbek Language

**B. Mansurov** and **A. Mansurov**
Copper City Labs
{b,a}mansurov@coppercitylabs.com

September 29, 2020

**Abstract**

In this paper, we share the process of developing word embeddings for the Cyrillic variant of the Uzbek language. The result of our work is the first publicly available set of word vectors trained on the word2vec, GloVe, and fastText algorithms using a high-quality web crawl corpus developed in-house. The developed word embeddings can be used in many natural language processing downstream tasks.

## 1 Introduction

In natural language processing (NLP) word embeddings are mappings of words to vectors of real numbers. They are used in many downstream tasks such as information retrieval [1], sentiment analysis [2], document summarization [3], and machine translation [4]. Pre-trained word embeddings for many languages are publicly available [5].

The Uzbek language is one of many low-resource languages, with very few publicly available data resources. Currently, both the Cyrillic and Latin scripts are used in the language. There are no known publicly available word embeddings for the Cyrillic variant of Uzbek. As far as we're aware, only fastText [5] word embeddings exist for the Latin variant. However, fastText was trained on the relatively low quality Uzbek Wikipedia and noisy Common Crawl corpus.

In this paper, we describe the process of generating three kinds of word embeddings (word2vec [6], GloVe [7], and fastText [8]) using high-quality data for the Cyrillic variant of the Uzbek language. We also publicly share our word vectors. Section 2 describes the process of gathering the training data. Section 3 provides the trained models and Section 4 shares the results. Section 5 summarizes our research and identifies areas for future research.

## 2 Data

To train our word embeddings, we crawled websites in the "uz" domain. For each website we manually created a scraper to extract only the article content written in Cyrillic Uzbek, and not other texts such as user interface elements or user comments.

We chose to parse Cyrillic texts because parsing texts written in the Latin script poses certain data issues. For example, the letter "ў" in Cyrillic can be written in many different forms in Latin: "o'",

"о`", "о´", "о ´", and "о ʼ". Although only one of them is correct, all of the above forms may appear in text, thus making it hard to correctly identify words. A similar issue happens with the letter "ғ" (g´).

The majority of websites we parsed are news websites because their archives are publicly available and they are continuously updated. In the future, if we want to release new versions of our models, we can simply retrieve new articles and create embeddings from the combined existing and new data. News articles also cover a wide range of topics. Websites we parsed include articles in the following categories: crime, culture, economics, health, politics, society, sports, and technology among others.

It's common for websites in Uzbek to provide content in Russian as well. To keep the training data clean, it was important to retrieve texts written in Uzbek only. Since Russian also uses the Cyrillic script, we made sure to ignore any texts containing even a single letter "ы" (which doesn't exist in Uzbek, but exists in Russian). This rule also ignores mainly Uzbek texts with a little bit of Russian included, but in order to keep the end result clean, we decided to accept this trade-off.

Once article contents were extracted, we lower-cased the words, and extracted tokens containing only the Cyrillic Uzbek letters and a hyphen. The total number of such tokens was 79,579,042.

We didn't parse Uzbek Wikipedia because most of its articles were automatically created from an Uzbek encyclopedia without human verification, and thus contains typos and abbreviations where usage of full words would be appropriate. Moreover, Uzbek Wikipedia is written in the Latin variant of Uzbek. Although Wikipedia provides a converter from Latin to Cyrillic, the end result contains many errors.

## 3 Training

We trained word2vec, GloVe, and fastText embeddings because they seem to be widely used by NLP practitioners. For example, the popular Python library Gensim[1] allows loading and using word2vec and fastText models. It is also possible to convert GloVe embeddings to word2vec format using the library. Another reason for the popularity of these algorithms is that their implementations are publicly available.

### 3.1 Word2vec

Word2vec was trained using the reference implementation[2]. The following table summarizes the trained models. The vocabulary generated by word2vec contains 366,276 unique tokens.

Table 1: Trained word2vec models. CBOW refers to continuous bag of words.

| Architecture | Training Algorithm | Word Vector Size | Window |
|---|---|---|---|
| CBOW | Negative Sampling | 100 | 5 |
| CBOW | Hierarchical Softmax | 300 | 5 |
| Skipgram | Negative Sampling | 100 | 10 |
| Skipgram | Hierarchical Softmax | 300 | 10 |

---

[1] https://radimrehurek.com/gensim/index.html
[2] https://github.com/tmikolov/word2vec

## 3.2 GloVe

We trained a 300-dimensional word vector model using the reference implementation[3].

## 3.3 FastText

FastText was also trained using the reference implementation[4]. The following table summarizes the trained models. In all models the minimum and maximum subword sizes were 2 and 5, respectively.

Table 2: Trained fastText models. CBOW refers to continuous bag of words.

| Architecture | Word Vector Size |
|---|---|
| CBOW | 100 |
| CBOW | 300 |
| Skipgram | 100 |
| Skipgram | 300 |

# 4 Results

Below are the CC BY 4.0 licensed **word embeddings** we share via Figshare.

- Word2vec 100 [9] and 300 [10] dimensional word vectors (CBOW, negative sampling);
- Word2vec 100 [11] and 300 [12] dimensional word vectors (skipgram, negative sampling);
- GloVe 300 dimensional word vectors [13];
- FastText 100 [14] and 300 [15] dimensional word vectors (CBOW);
- FastText 100 [16] and 300 [17] dimensional word vectors (skipgram).

The shared data also include the hyper-parameters used in training.

Next, we present some snippets from the training data, followed by nearest neighbor examples from the model outputs. The table below shows the most and least frequently appearing words in the training data.

Table 3: Most and least frequently appearing words in the training data.

| Word | Frequency | Word | Frequency |
|---|---|---|---|
| ва (and) | 1550236 | ликсеич (Likseich) | 5 |
| билан (with) | 767708 | жамбатиста (Giambattista) | 5 |
| ҳам (also) | 699727 | глазичевдан (from Glazevich) | 5 |
| бу (this) | 606904 | жамиятшуносга (to sociologist) | 5 |
| бир (one) | 544584 | биг-бэнд (big band) | 5 |
| учун (for) | 493722 | майиб-муғтало (utterly disabled) | 5 |
| шу (this) | 274099 | ота-турна (father crane) | 5 |
| эди (was) | 229156 | нилкантога (to Nilkanto) | 5 |
| деб (saying) | 228540 | кингисепп (Kingisepp) | 5 |
| ўз (self) | 215465 | рухсатой (Ruxsatoy) | 5 |

---

[3]https://github.com/stanfordnlp/GloVe
[4]https://github.com/facebookresearch/fastText

For our nearest neighbor examples, we chose a frequent and a rare word from the corpus. In the corpus the word сув (water) appeared 40,300 times (frequent word), while the word нордон (sour) appeared 200 times (rare word). The ten nearest neighbors for each word and model are given below. An English translation of each word is given in parentheses. Repeating translations (e.g., water) are primarily due the morphological richness of the Uzbek language.

- 100-dimensional word2vec model (CBOW, Negative Sampling)
  - **сув**: сувни (water), сувларни (waters), коллектор (water collector), водопровод (water pipes), суви (water), сувнинг (of water), сувини (water), сувларнинг (of waters), сувлар (waters), сувлари (waters)
  - **нордон**: таъмли (tasteful), шакарли (sweet), тахир (bitter), таъми (taste), ялпизли (mint), хушхўр (pleasant), шираси (juice), хуштаъм (delicious), шакари (sugar), шарбатларни (juices)

- 300-dimensional word2vec model (CBOW, Hierarchical Softmax)
  - **сув**: сувни (water), сувлар (waters), сувнинг (of water), сувини (water), сувларни (waters), сувдан (from water), сувлари (waters), суви (water), сувли (watery), лойқа (muddy)
  - **нордон**: мазали (tasty), ёғли (greasy), ширин (sweet), аччиқ (bitter), бемаза (bland), хушбўй (fragrant), тансиқ (yummy), ейиш (eating), сархил (flavorful), қуритилган (dried)

- 100-dimensional word2vec model (Skipgram, Negative Sampling)
  - **сув**: сувни (water), суви (water), сувнинг (of water), сувдан (from water), сувини (water), лойқа (muddy), сувлари (waters), сувга (water), сувларни (waters), сувига (water)
  - **нордон**: таъмли (tasteful), шарбатлар (juices), цитруслилар (citrus), ўпка-жигар (lung-liver), қовурилган (fried), дудланган (smoked), шираси (juice), апельсин (orange), таъм (taste), қовурма (fried food)

- 300-dimensional word2vec model (Skipgram, Hierarchical Softmax)
  - **сув**: сувни (water), сувнинг (of water), лойқа (muddy), сувга (water), сувдан (from water), сувини (water), суви (water), сувлар (waters), сувлари (waters), сувига (water)
  - **нордон**: таъмли (tasteful), апельсин (orange), шарбатлар (juices), егуликлар (foods), шираси (juice), ёғли (greasy), шарбат (juice), мева (fruit), мандарин (mandarin), газланган (carbonated)

- 300-dimensional GloVe model
  - **сув**: суви (water), сувни (water), ичимлик (drink), тупроқ (soil), сувдан (from water), тоза (clean), ер (land), сувга (water), газ (gas), табиий (natural)
  - **нордон**: таъмли (tasteful), таъми (taste), егуликлар (foods), болдан (honey), тахир (bitter), кифояли (enough), қалампирнинг (of pepper), болдек (like honey), шўр (saline), мазали (tasty)

- 100-dimensional fastText model (CBOW)
  - **сув**: сув-сув (water-water), сув-пув (water-drinks), сувлиқ (watery), сувд (water), сувлоқ (watery), сувлуқ (wateriness), сувин (have water), сувдон (jug), суву (water), сувот (watering)
  - **нордон**: нордон-шўр (sour-salty), гордон (gordon), ордон (sour like), кордон (cordon), нордонгина (soury), нортон (norton), гардон (keeper), нор (markedly), норд (sour), норчучук (sour like)

- 300-dimensional fastText model (CBOW)

- **сув**: сув-сув (water-water), сув-пув (water-drinks), сувд (water), суву (water and), сувлиқ (watery), сувлуқ (wateriness), сувлоқ (watery), сувин (have water), сувлисой (creek), сувг (water)
- **нордон**: нордон-шўр (sour-salty), ордон (sour like), гордон (gordon), норд (sour), кордон (cordon), нордонгина (soury), нортон (norton), нор (markedly), нордонроқ (sour like taste), норча (soury taste)

- 100-dimensional fastText model (Skipgram)
  - **сув**: сувни (water), суви (water), сувин (have water), сувлар (waters), сувини (water), сувларни (waters), сувлаб (watering), сув-сув (water-water), сувим (my water), сувд (water)
  - **нордон**: нордон-шўр (sour-salty), нордонроқ (sour like taste), нордонгина (soury), таъмли (tasteful), ошқошиқ (spoon), ананас (pineapple), ачқимтил (bitterly), кефир (kefir), апельсин (orange), шираси (juice)

- 300-dimensional fastText model (Skipgram)
  - **сув**: сувни (water), суви (water), сувин (have water), сувд (water), сув-сув (water-water), сувнинг (of water), сувим (my water), сувини (water), сувлар (waters), сувдан (from water)
  - **нордон**: нордон-шўр (sour-salty), нордонгина (soury), нордонроқ (sour like taste), таъмли (tasteful), бордон (mour), норд (sour), ордон (sour), кефир (kefir), шакарсиз (sugarless), шакарли (sugary)

# 5 Conclusion

We developed and shared multiple word embedding models for the Cyrillic variant of the low-resource Uzbek language. Our hope is that these models will be useful in research and real-world applications.

As part of this work we didn't analyze the quality and performance of our trained word vectors. We leave it as an exercise for future work. In the future we'd also like to train word embeddings using various sources of data, not just websites.

Another improvement is related to correctly identifying texts written in Uzbek. Our heuristic approach of ignoring Russian articles may have worked well in this case, but in the future, with a diverse set of training data, we will need to employ more robust methods of identifying and excluding non-Uzbek texts from our training data set.

Word vectors that we generated are context-independent — they output only one vector per word. Creating context-dependent word embeddings such as generated by BERT [18] and ELMo [19] is another area to explore.